\pdfoutput=1

\documentclass[11pt]{article}

\usepackage{acl}

\usepackage{times}
\usepackage{latexsym}
\usepackage{lipsum}
\usepackage{multirow}
\usepackage{booktabs}
\usepackage{pdflscape}
\usepackage{soul,color}
\usepackage{graphicx}
\usepackage[export]{adjustbox}
\usepackage{caption}
\usepackage{subcaption}
\usepackage{amsmath}
\usepackage[edges]{forest}
\usepackage{enumerate}

\usepackage[edges]{forest}
\usepackage{caption}
    \tikzset{parent/.style={align=center,text width=3cm,rounded corners=3pt},
        child/.style={align=center,text width=3cm,rounded corners=3pt}
    }

    \colorlet{col0}{violet!20}
    
    \colorlet{col1}{red!20}
    \colorlet{col2}{blue!20}
    \colorlet{col3}{green!20}

\usepackage[T1]{fontenc}
\usepackage[utf8]{inputenc}

\usepackage{microtype}

\usepackage{inconsolata}

\newcommand{\midad}{\textit{Midad}}
\title{\textit{Qalam}\protect\includegraphics[height=1cm]{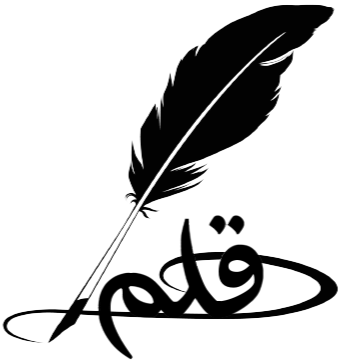}: A Multimodal LLM for Arabic Optical Character and Handwriting Recognition}

\author{
\textbf{Gagan Bhatia} ~~~~
\textbf{El Moatez Billah Nagoudi} ~~~~ \\ \textbf{Fakhraddin Alwajih} ~~~~\textbf{Muhammad Abdul-Mageed}
\\The University of British Columbia \& Invertible AI \\
\texttt{\normalsize \{gagan30@student.,muhammad.mageed@\}ubc.ca} \\  
}

\begin{document}
\maketitle
\begin{abstract}
Arabic Optical Character Recognition (OCR) and Handwriting Recognition (HWR) pose unique challenges due to the cursive and context-sensitive nature of the Arabic script. This study introduces \textit{Qalam}, a novel foundation model designed for Arabic OCR and HWR, built on a SwinV2 encoder and RoBERTa decoder architecture. Our model significantly outperforms existing methods, achieving a Word Error Rate (WER) of just 0.80\% in HWR tasks and 1.18\% in OCR tasks. We train \textit{Qalam} on a diverse dataset, including over 4.5 million images from Arabic manuscripts and a synthetic dataset comprising 60k image-text pairs. Notably, \textit{Qalam} demonstrates exceptional handling of Arabic diacritics, a critical feature in Arabic scripts. Furthermore, it shows a remarkable ability to process high-resolution inputs, addressing a common limitation in current OCR systems. These advancements underscore \textit{Qalam}'s potential as a leading solution for Arabic script recognition, offering a significant leap in accuracy and efficiency.
\end{abstract}

\section{Introduction}\label{intro}
Optical Character Recognition (OCR) technology has revolutionized the way we interact with written and printed materials. It enables the conversion of various documents, including scanned paper documents, PDF files, or images captured by a digital camera, into editable and searchable data. The ability of OCR to digitize text has found applications in numerous domains, ranging from banking and healthcare to education and historical research, among others \cite{singh2012survey}.



\begin{figure}[!t]
\includegraphics[width=\columnwidth]{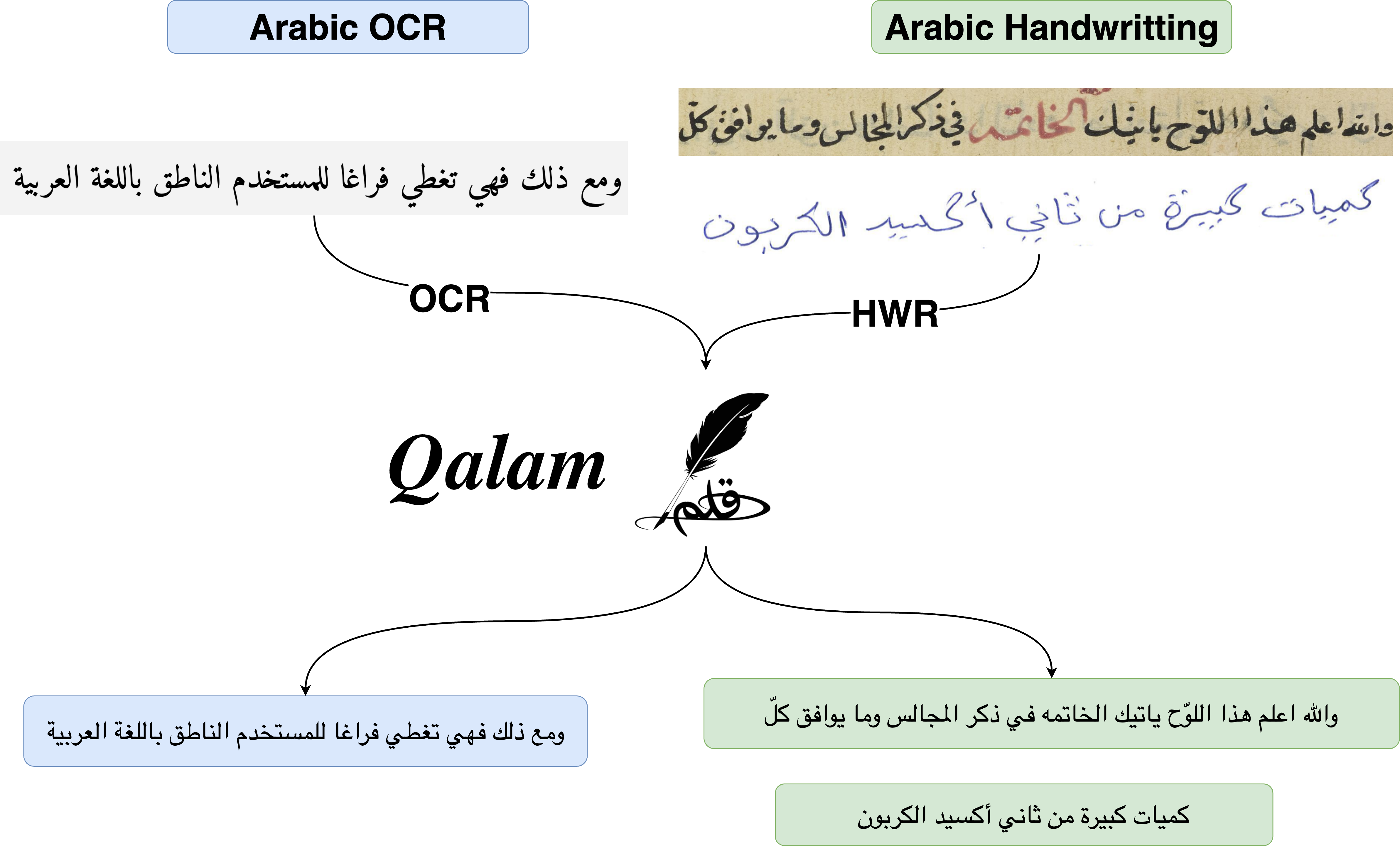}
\caption{An illustrative overview of \textit{Qalam}'s functioning for Arabic OCR and HWR across diverse text types.}
\label{fig:shapes}
\end{figure}

\begin{figure}[t!]
\includegraphics[width=\linewidth]{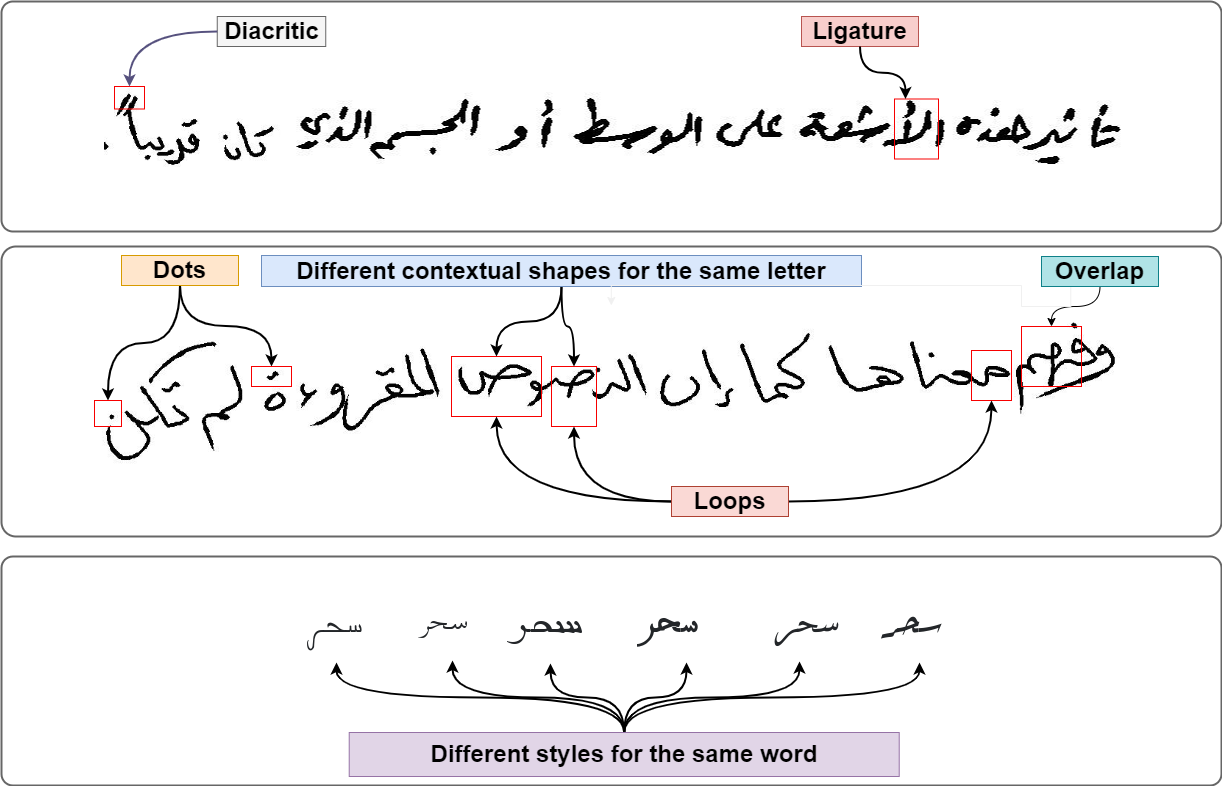}
\caption{An illustrative depiction highlighting the distinctive characteristics of Arabic script that contribute to its increased complexity compared to other languages.}
\label{fig:challenges}
\end{figure}

In this work, our focus is on handling Arabic OCR and HWR. Arabic OCR and HWR pose substantial challenges due to several distinctive features of the Arabic script. The Arabic writing system is cursive and context-dependent, a characteristic that complicates the design of robust OCR models. Further intricacies such as diacritical marks, loops, overlaps, ligatures, dots, and multi-context joining contribute to the complexity of the task.  In addition to these structural challenges, the vast diversity of Arabic fonts and individual handwriting styles further complicate character recognition and segmentation tasks. Finally, the lack of comprehensive and diverse annotated datasets increases the difficulty of Arabic OCR and HWR tasks. Figure \ref{fig:challenges} visually represents these challenges.

OCR systems can be classified based on the type of input data they process: handwritten or printed. In this paper, for clarity, we refer to printed text recognition specifically as OCR, although both types broadly fall under the OCR umbrella.

Our objective in this work is to tackle these complex tasks, making several key contributions that collectively underscore the novelty of our research. Concretely, our contributions can be summarized as follows:

\begin{enumerate}[(i)]
\item We introduce \textit{Qalam}, a novel OCR and HWR model specifically engineered for Arabic script, establishing a new state-of-the-art performance across diverse datasets.
\item We compile a large and diverse collection of datasets and introduce \midad~Benchmark for Arabic OCR and HWR. This will serve as a resource for future research in the community.
\item Our work offers an in-depth analysis of the inherent complexities associated with Arabic script OCR, contributing towards a broader understanding of the challenges in this domain.
\item Last, we provide detailed comparative evaluations against multiple baselines, thus emphasizing the effectiveness of our approach.
\end{enumerate}

The rest of this paper is structured as follows: Section \ref{RW} discusses the related work in HWR and OCR. Section \ref{datasets} introduces \midad~, our Arabic HWR and OCR benchmark. In Section \ref{method}, we present our methodology. Section \ref{experiments} describes the experiments conducted in this work. Section \ref{qalam} introduces \textit{Qalam}, the proposed model for Arabic HWR and OCR, and evaluates \textit{Qalam}. We conclude in Section \ref{conc} and spell out many limitations related to our work in Section \ref{lim}.

\section{Related Works}\label{RW}

Traditional Hidden Markov Model (HMM) \citep{bunke1995off,park1996off, agazzi1993hidden} approaches in sequence modeling have been largely surpassed by deep learning techniques that do not require explicit segmentation. Connectionist Temporal Classification (CTC) models \citep{graves2006connectionist,graves2008offline} and Encoder-Decoder models \citep{sutskever2014sequence,bluche2017scan} with attention mechanisms \citep{bahdanau2014neural,michael2019evaluating} represent two primary deep learning categories. Recent advances include transformer models \citep{vaswani2017attention} and pre-trained models \citep{devlin2018bert}.

\paragraph{Handwriting Recognition (HWR).} HMMs were traditionally used for HWR \citep{bunke1995off,park1996off}, but CTC-based models \citep{graves2006connectionist,graves2008offline} became more popular due to their accuracy without explicit segmentation. These models often use RNNs and their variations like LSTM \citep{pham2014dropout}, BLSTM, and MLSTM \citep{graves2008offline,voigtlaender2016handwriting,bluche2017scan}, and combinations with CNNs such as CNN-BLSTM \citep{de2020htr}. Recurrence-free models optimized with CTC, using only CNNs, have also been developed \citep{coquenet2020recurrence}. 

\paragraph{Optical Character Recognition (OCR).} OCR, divided into Scene Text Recognition (STR), Scanned Document OCR, and Synthetic Text OCR, has evolved from HMMs to pre-trained transformer models. RNN and CTC-based models \citep{su2015accurate}, combined CNN and BLSTM models \citep{shi2016end,breuel2017high}, and Encoder-Decoder architectures with attention mechanisms \citep{lee2016recursive,shi2018aster} have been used. Recently, transformer-based models \citep{li2021trocr,kim2022ocr,lyu2022maskocr} have gained prominence.

\paragraph{Multimodal Large Language Models (MLLMs).} MLLMs have impacted tasks like Visual Question Answering (VQA) and OCR-related tasks \citep{zhang2024vision,wadhawan2024contextual,shi2023exploring}. Despite their success, challenges in text recognition within images remain due to lower encoder resolution \citep{liu2023hidden}. Open-source models like LLaVAR \citep{zhang2023llavar} and MonkeyText \citep{liu2024textmonkey} are improving text recognition, while closed-source models like GPT-4V \citep{achiam2023gpt} and Gemini Pro \citep{team2023gemini} contribute to these advancements.

\paragraph{Arabic HWR and OCR.} Arabic HWR and OCR initially relied on HMMs \citep{alma2002recognition,prasad2008improvements}. Later, CTC-based models with RNNs and CNNs replaced HMMs. \citet{ahmad2017khatt} used an MDLSTM model for the KHATT dataset, and \citet{yousef2020accurate} introduced a pure CNN model optimized with CTC loss. Recent advancements include transformer models \citep{mostafa2021ocformer,momeni2023transformer}, though Arabic MLLMs still face challenges in text recognition within images \citep{alwajih2024peacock}. For more detailed surveys, see \citet{alrobah2022arabic,faizullah2023survey} and reviews by \citet{sobhi2020arabic,alghyaline2023arabic}. Figure \ref{fig:classification} (Appendix \ref{models_arch}) categorizes various model architectures used in these studies.


\section{\textit{MIDAD} Benchmark} \label{datasets}

\subsection{Datasets}
We utilize a variety of printed and handwritten Arabic OCR datasets in this study, which we have collectively named \textit{MIDAD}. Below is a summary of these datasets.

\noindent\textbf{MADBase.} A database of $60k$ training and $10k$ testing images of Arabic handwritten digits. It's often used for training and testing CNNs \cite{MADBase}.

\noindent\textbf{AHCD.} Similar to MADBase, the AHCD dataset is used for training and testing CNNs, but contains $16$k samples of handwritten Arabic characters \cite{AHCD}. 

\noindent\textbf{ADAB.} Consists of $937$ Tunisian town and village names in Arabic handwriting \cite{ADAB2}. It contains 15k samples in total.

\noindent\textbf{Alexuw.} This dataset is compiled by \newcite{alexuw}, includes $25k$ samples of 109 different Arabic words. This large and diverse dataset allows for the development and testing of segmented letter-based Arabic handwriting recognition algorithms.

\noindent\textbf{Shotor.} Introduced by~\newcite{shotor}, this is a large-scale open-source dataset, consisting of $120k$ grayscale images of various Farsi phrases, each rendered in different fonts and sizes. The phrases were sourced from the Ganjoor and Farsi Wikipedia websites~\cite{shotor}.

\noindent\textbf{PATS-A01.}
Introduced by~\newcite{PATS}, it represents the first printed Arabic text set containing $22k$ line images sourced from classic Arabic literature. Eight different Arabic fonts were used for these images, adding to the variety in this dataset \cite{PATS}.

\noindent\textbf{IDPL-PFOD.}
This synthetic dataset, created by\newcite{IDPL}, includes $30k$ TIF images of printed Farsi text lines. Each image has varying background types and levels of blur and distortion to mimic real-life conditions. The text is rendered using popular Farsi typefaces in diverse sizes and styles.

\noindent\textbf{UPTI.}
The UPTI dataset was developed by \newcite{sabbour} and contains $10k$ synthetic Urdu text lines in the Nastaliq font, providing a valuable resource for testing and training models in a language closely related to Arabic \cite{UPTI}.

\noindent\textbf{OnlineKHATT.} A comprehensive dataset consisting of $10k$ Arabic text lines, created by $623$ authors using various devices. The data come as stored in the online format using InkML files, and we convert it into offline format as images. The dataset provides character-based segments and manually verified ground truths for the written lines \cite{mahmoud2018online}.
\textit{MIDAD} provides a broad and varied base for training and validating OCR models for Arabic and other closely related languages.

\begin{table*}[]
\centering
\resizebox{\linewidth}{!}{%
\begin{tabular}{lrrrrrrrrrrrr}\toprule
\textbf{Dataset} &\textbf{Citation} &\textbf{Base} &\textbf{Type} &\textbf{Samples} &\textbf{Language} &\textbf{Dimensions} &\textbf{Sentence Length} &\textbf{Words} &\textbf{Train} &\textbf{Dev} &\textbf{Test} \\\midrule
MADBase &\newcite{MADBase} &\multirow{2}{*}{Char} &HW &$70k$ &Arabic &(28,28) &1 &10 &48,000 &6,000 &6,000 \\
AHCD &\newcite{el2017arabic} & &HW &$16k$ &Arabic &(28,28) &1 &10 &10,752 &1,344 &1,344 \\
ADAB &\newcite{el2009icdar} &\multirow{3}{*}{Word} &HW &$15k$ &Arabic &(300,80) &1.53 &960 &12,022 &1,503 &1,503 \\
Alexuw &\newcite{hussein2014alexu} & &HW &$25k$ &Arabic &(480,232) &1 &10,989 &20,091 &2,512 &2,511 \\
Shotor &\newcite{shotor} & &OCR &$120k$ &Persian &(100,50) &1.08 &62,900 &96,000 &12,000 &12,000 \\
PATS01 &\newcite{PATS} &\multirow{3}{*}{Line} &OCR &$22k$ &Arabic &(1344,80) &16.65 &8,248 &17,702 &2,213 &2,213 \\
IDPL-PFOD &\newcite{sadat2021idpl} & &OCR &$30k$ &Persian &(700,50) &15 &38,476 &24,110 &3,014 &3,014 \\
OnlineKHATT &\newcite{mahmoud2018online} & &HW &$10k$ &Arabic &(1089,150) &7.42 &22,216 &6,798 &850 &850 \\
\bottomrule
\end{tabular}}
\caption{Summary of Dataset Characteristics: The table presents a comprehensive overview of the datasets used in the study. These datasets collectively offer a diverse range of samples, ensuring the robustness and adaptability of the models evaluated.}
\label{tab:datasets}
\end{table*}


\subsection{Data Splits}
In instances where the original datasets came pre-partitioned into standard training (Train), development (Dev), and testing (Test) splits, we opt to maintain these existing divisions to preserve the integrity of the original data structure. In absence of such pre-defined splits, we randomly shuffle each dataset and split it into three 80\% Train, 10\% Dev, and 10\% Test.  Comprehensive statistics reflecting the distribution of data splits across all our datasets are in Table~\ref{tab:datasets}.

\subsection{In the wild Arabic OCR and HWR Datasets}

\begin{figure}[htbp]
     \centering
     \begin{subfigure}[b]{\columnwidth}
         \centering
         \includegraphics[width=\columnwidth]{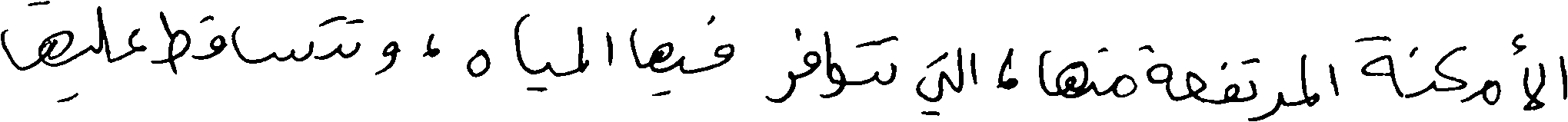}
     \end{subfigure}
     \hfill
     \begin{subfigure}[b]{\columnwidth}
         \centering
         \includegraphics[width=\columnwidth]{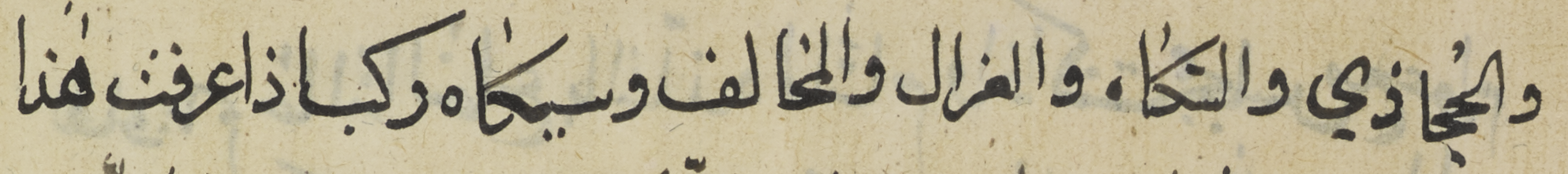}
     \end{subfigure}
     \hfill
        \caption{In-the-wild Arabic dataset samples.}
        \label{fig:khatt}
\end{figure}

\noindent\textbf{KHATT} The KHATT dataset \cite{articlekhatt}, a prominent tool in Arabic handwriting recognition, offers $1,000$ handwritten Arabic forms from diverse writers, totalling $4,000$ paragraph images—half with similar text and the other half with unique text. Our study concentrated on the unique-text paragraphs providing $6,742$ text lines post-segmentation.

\noindent\textbf{Historical Manuscripts} The dataset contains $120$ images from historical Arabic manuscripts, digitized by the British Library and Qatar Foundation \cite{clausner2018icfhr}.

\begin{figure}[h]
\centering
\includegraphics[width=\columnwidth]{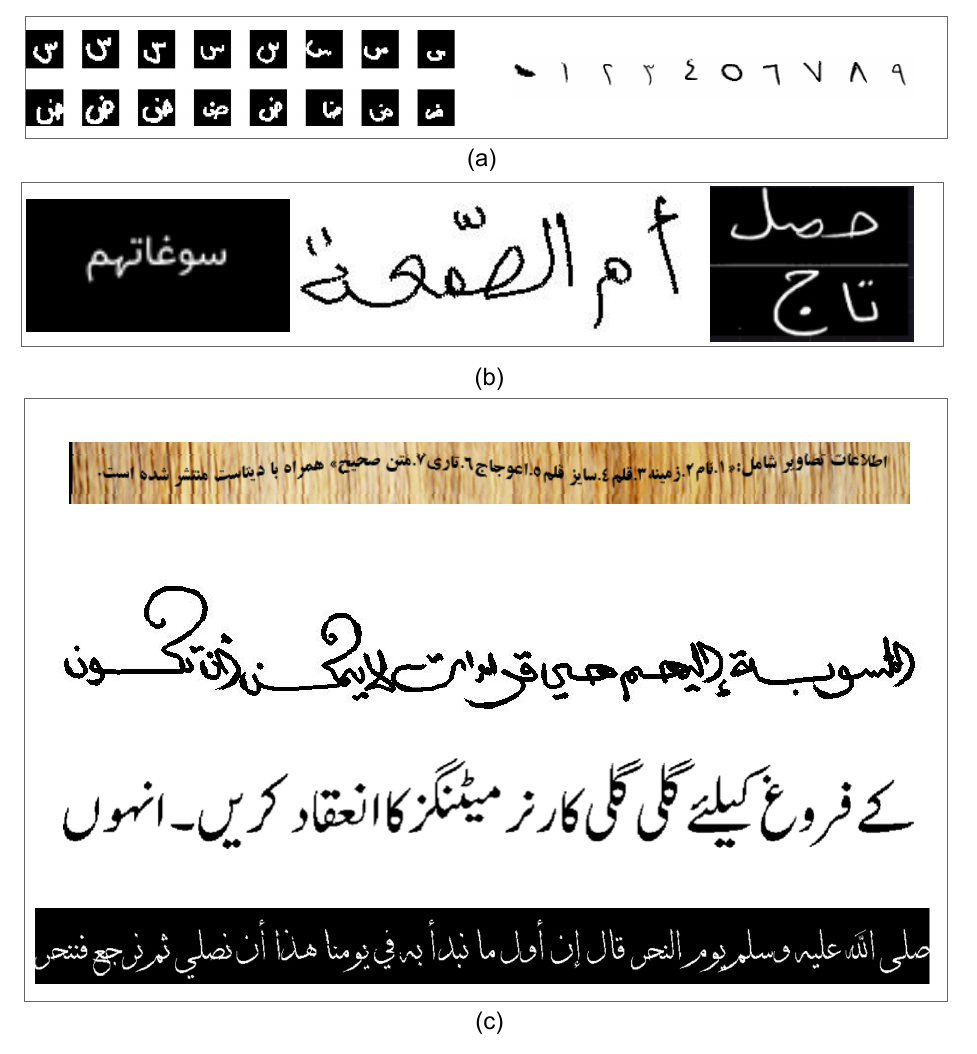}
\caption{ A showcase of diverse Arabic script datasets, illustrating the intricate and multifaceted challenges addressed by \textit{Qalam}: (a) Character-based examples, (b) Word-oriented examples, (c) Line-based examples}
\label{fig:shapes}
\end{figure}

\section{Methods}  \label{method}

In our pursuit to address the challenges of both Arabic HWR and OCR, we employ a Vision Encoder-Decoder (VED) framework that ingeniously brings together transformer-based models in a novel manner. This framework leverages the power of transformer-based vision models as encoders, adeptly processing image data, and pairs it with the linguistic sophistication of transformer-based language models as decoders. The result is a synergistic pairing that skillfully transcodes visual information into meaningful textual output, thus overcoming the intricate complexities of Arabic OCR. Furthermore, we extensively analyze various encoder and decoder combinations, investigating their implications for model performance. In VED design, the encoder ingests image data, while the decoder manages ground-truth caption inputs via teacher forcing \cite{sutskever2014sequence} during training. We ensure autoregressive training for the next token prediction using causal self-attention in the model. This mechanism restricts a token's attention only to its predecessors, maintaining the sequential nature of the input text.


\subsection{Encoder Configuration}

Our encoder takes images resized to $(H,W)$, padded for uniformity, and partitioned into $N = H*W/P^2$ patches with a fixed size of $(P,P)$. These patches are subsequently flattened and linearly projected into $D$-dimensional vectors to form patch embeddings. We retain the $``[CLS]"$ token to represent the image and incorporate learnable 1D position embeddings based on absolute positions.




\subsection{Decoder Configuration}

The decoder in our VED structure consists of layers identical to the encoder, but with an additional \textit{encoder-decoder attention} mechanism. It employs attention masking to prevent the model from peeping into the future during training. Its hidden states are projected linearly to the vocabulary size, and probabilities are computed using softmax. We initialize the cross-attention layer weights randomly when warm-starting from pre-trained transformer-based models.


\subsection{Baselines}
To provide a comparative analysis, we consider a range of established OCR models as baselines. These models, varying in complexity and technique, include CRNN, Gated-CNN-BiLSTM-CTC, Tesseract, and TrOCR.

\noindent\textbf{CRNN.} The CRNN model \citep{3DRNN} blends convolutional and recurrent layers, and concludes with a linear layer and CTC loss function. It leverages dropout, batch normalization, LeakyReLU, Maxpool, and bidirectional 1D-LSTM layers. Data augmentation through random distortions is used for enhanced robustness.

\noindent\textbf{Gated-CNN-BiLSTM-CTC.} This architecture \citep{bluche2017gated} includes Gated-CNNs for feature extraction. Convolutional layers, gates, bidirectional LSTM layers, and a linear layer compose the encoder-decoder structure. The model minimizes the CTC objective function using RMSProp. It also incorporates data augmentation.

\noindent\textbf{Tesseract.} Introduced by~\newcite{patel}, this is a versatile open-source OCR engine sponsored by Google. It is designed to be a universal text recognition tool, recognizing over 100 languages. Tesseract works by decomposing the input image into lines and words.

\noindent\textbf{TrOCR.} Introduced by~\newcite{li2021trocr}, this is a text recognition model that uses pre-trained image and text transformer-based models. Pre-training on large-scale synthetic data, and fine-tuning on human-labeled datasets make TrOCR effective in printed, handwritten, and scene text recognition tasks.

\subsection{Evaluation Metrics}

Model performance is evaluated based on the Word Error Rate (WER), computed as the normalized Levenshtein distance at the word level. The formula to calculate WER is given in Appendix \ref{wer_eq}.

\section{Experiments} \label{experiments}
We present our experimental settings, including the selection process for the encoder and decoder from various options. We also discuss the challenges encountered during their selection.


\subsection{Encoder Selection}

The second phase of our experimental design involved selecting the most suitable encoder from available options. We conducted rigorous tests with four different encoders (ViT \cite{dosovitskiy2020image}, DeiT \cite{DEIT}, BEiT \cite{bao2021beit}, Swin \cite{swin}, and SwinV2 \cite{liu2022swin} Transformers) while keeping the XLM-RoBERTa \cite{roberta} as the constant decoder. To ensure the uniformity of these experiments, we used the hyperparameters derived from the initial tuning stage.

\noindent{\textbf{Results.}} Table \ref{tab: encoder_table} displays a performance comparison of several models, including ViT, Swin, BeiT, SwinV2, and DeiT using a constant decoder XLM-R on OCR and HWR tasks. Overall, the DeiT encoder model exhibits superior results and it has a \midad~score of 19.79, likely due to its adeptness at the discerning text in various forms and orientations in images and recognizing intricate handwritten patterns. However, the SwinV2 encoder model which has a \midad~score of 21.60 also shows notable performance, particularly inline-level handwriting recognition on the `OnlineKHATT' dataset, due to its ability to handle variable image inputs Figure \ref{fig:high_res_example} explains this visually. The results underline the effectiveness of both DeiT and SwinV2 in handling diverse OCR and HWR tasks across \textit{Midad} datasets.

\begin{table}[!htp]\centering
 \scriptsize
\resizebox{\linewidth}{!}{%
\begin{tabular}{lrrrrrrrr}\toprule
\textbf{Cluster} &\textbf{Task} &\textbf{Dataset} &\textbf{\colorbox{green!25}{E1}} &\textbf{\colorbox{red!25}{E2}} &\textbf{\colorbox{orange!25}{E3}} &\textbf{\colorbox{blue!25}{E4}} &\textbf{\colorbox{purple!25}{E5}} \\\midrule
\multirow{5}{*}{\textbf{HWR}} &Char &MADBase &7.49 &3.39 &6.84 &3.40 &\textbf{2.74} \\
& &AHCD &10.00 &8.84 &5.95 &5.21 &\textbf{3.73} \\
&Word &ADAB &6.92 &3.02 &8.92 &3.05 &\textbf{2.85} \\
& &Alexuw &13.92 &8.07 &7.78 &7.92 &\textbf{4.78} \\
&Line &OnlineKHATT &75.02 &66.39 &65.23 &\textbf{64.10} &65.03 \\\midrule
\multirow{3}{*}{\textbf{OCR}} &Line &PATS01 &52.73 &37.74 &41.73 &\textbf{33.13} &35.24 \\
&Word &Shotor &11.74 &8.30 &7.73 &7.66 &\textbf{7.03} \\
&Line &IDPL-PFOD &46.84 &48.43 &31.39 &30.20 &\textbf{29.42} \\\midrule
\multirow{3}{*}{\textbf{Overall}} &\multirow{3}{*}{\textbf{Avg.}} &HWR Score &22.67 &17.94 &18.94 &16.74 &\textbf{15.83} \\
& &OCR Score &37.10 &31.49 &26.95 &23.66 &\textbf{23.90} \\
& &\colorbox{yellow}{\textbf{\textit{\midad~\textsubscript{Score}}}}
&28.08 &23.02 &21.95 &19.33 &\textbf{18.85} \\
\bottomrule
\end{tabular}}
\caption{Comparative performance analysis using WER (lower is better) of various models across diverse OCR and HWR datasets on the validation set. \colorbox{green!25}{E1}: ViT, \colorbox{red!25}{E2}: Swin, \colorbox{orange!25}{E3}: BeiT, 
\colorbox{blue!25}{E4}: SwinV2, 
and \colorbox{purple!25}{E5}:DeiT with XLM-R as decoder. The table also provides OCR, HWR, and \textit{Midad} scores to showcase the models' overall performance in respective tasks.}\label{tab: encoder_table_dev}
\end{table}

\subsection{Decoder Selection}

With DeiT established as the optimal encoder, we proceeded to the third phase where we tested the efficacy of different decoders. Maintaining DeiT as a constant encoder, we experimented with five different decoders, assessing their performance on all datasets under the same hyperparameters. 
We used RoBERTa \cite{liu2019roberta}, XLM-R \cite{roberta}, ARBERT, MARBERT and MARBERTv2 \cite{arbert}.

\noindent{\textbf{Results.}} Table~\ref{tab: decoder_table} contrasts the performance of transformer-based decoders, we have fixed the encoder to DeiT and are experimenting with different decoders on various HWR and OCR tasks. Generally, ARBERT stands out with the lowest WER across most tasks and datasets, showing its strong deciphering ability for diverse texts and handwriting. It acquires a \midad~score of 12.06 WER for HWR and 18.83 for OCR, with an overall (on both tasks) \midad of 17.02 WER.  Marbert\textsubscript{v2}, however, excels specifically on the OCR `IDPL-PFOD' which is a Persian line-level task, implying its potential suitability for specific OCR scenarios with dataset which have different languages that use the Arabic Script. These findings highlight ARBERT's overall dominance, while suggesting the usefulness of MARBERT\textsubscript{v2} for particular tasks, emphasizing the importance of task-specific model selection.

\begin{table}[!htp]\centering
\scriptsize
\resizebox{\linewidth}{!}{%
\begin{tabular}{lrrrrrrrr}\toprule
\textbf{Cluster} &\textbf{Task} &\textbf{Dataset} &\textbf{\colorbox{green!25}{D1}} &\textbf{\colorbox{red!25}{D2}} &\textbf{\colorbox{orange!25}{D3}} &\textbf{\colorbox{blue!25}{D4}} &\textbf{\colorbox{purple!25}{D5}} \\\midrule
\multirow{5}{*}{\textbf{HWR}} &Char &MADBase &$13.03$ &3.74 &6.03 &2.15 &\textbf{0.52} \\
& &AHCD &14.49 &4.73 &9.6 &2.81 &\textbf{1.85} \\
&Word &ADAB &9.51 &3.85 &6.57 &3.03 &\textbf{1.13} \\
& &Alexuw &13.92 &5.78 &9.74 &5.43 &\textbf{1.93} \\
&Line &OnlineKHATT &75.26 &66.03 &67.59 &58.73 &\textbf{54.84} \\\midrule
\multirow{3}{*}{\textbf{OCR}} &Line &PATS01 &54.03 &31.71 &45.61 &27.58 &\textbf{22.92} \\
&Word &Shotor &11.39 &8.03 &7.04 &5.51 &\textbf{3.62} \\
&Line &IDPL-PFOD &64.6 &30.42 &36.15 &\textbf{27.92} &28.45 \\\midrule
\multirow{3}{*}{\textbf{HWR+OCR}} &\multirow{3}{*}{\textbf{Avg.}} &HWR Score &28.30 &16.83 &19.91 &14.43 &\textbf{12.05} \\
& &OCR Score &43.34 &23.39 &29.60 &20.34 &\textbf{18.33} \\
& &\colorbox{yellow}{\textbf{\textit{\midad~\textsubscript{Score}}}} &34.74 &19.29 &23.54 &16.64 &\textbf{14.41} \\
\bottomrule
\end{tabular}}
\caption{Performance comparison using WER (lower is better) of various transformer-based decoder models on HWR and OCR tasks for different datasets on the validation set. \colorbox{green!25}{D1}: RoBERTa, \colorbox{red!25}{D2}: XLM-R, \colorbox{orange!25}{D3}: MARBERT, \colorbox{blue!25}{D4}: MARBERT\textsubscript{v2}, and \colorbox{purple!25}{D5}: ARBERT with DeiT as constant encoder. Tasks are categorized into character-level (Char), word-level (Word), and line-level (Line) recognition.}\label{tab: decoder_table_dev}
\end{table}

\subsection{Error Analysis}
\begin{figure}[htp]
\centering
\includegraphics[width=\columnwidth]{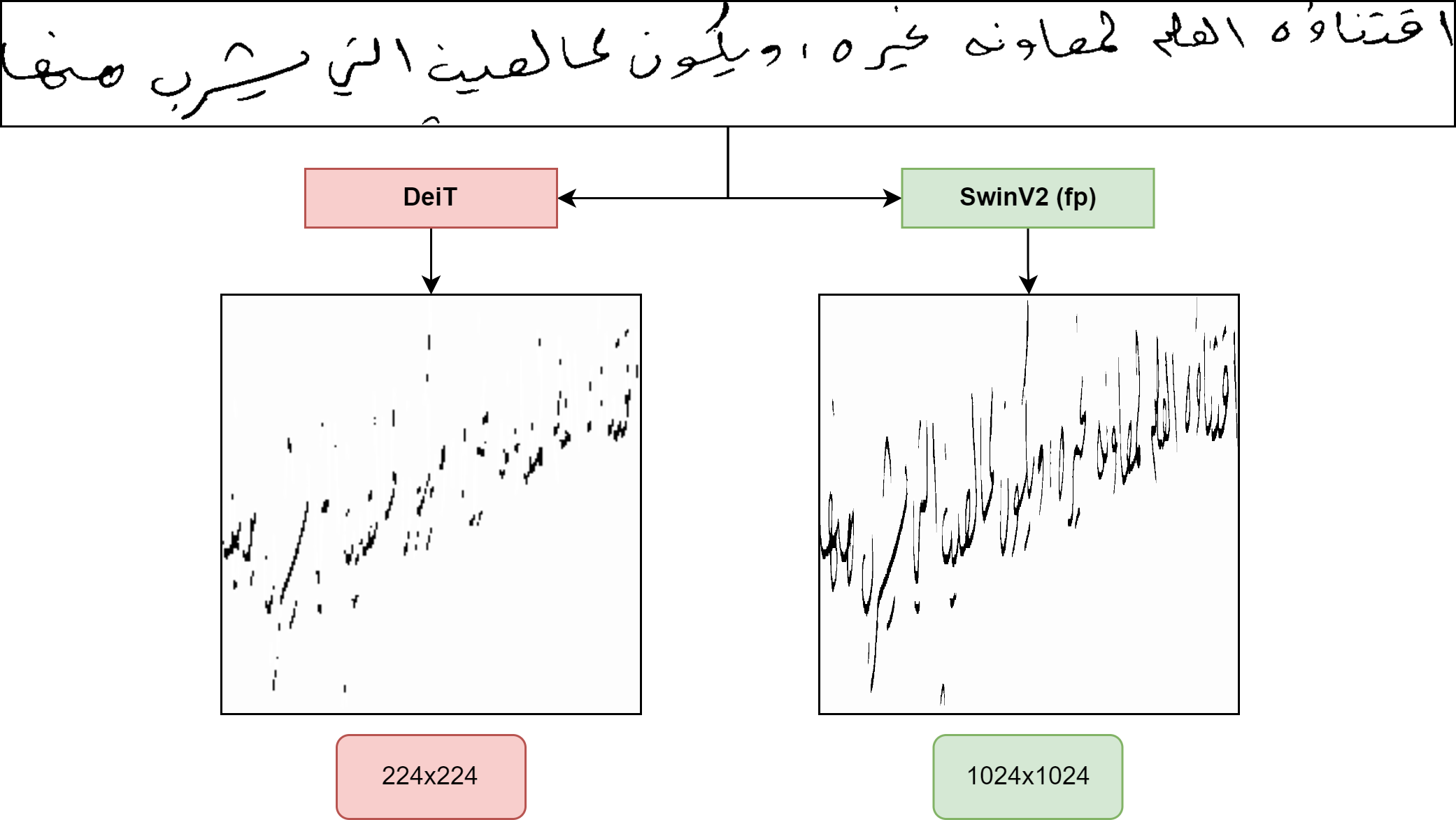}
\caption{Example of a high-resolution image that may present processing difficulties for DeiT. The images are reconstructions produced by both DeiT and SwinV2 following MIM training.}
\label{fig:high_res_example}
\end{figure}

We carry out a manual error analysis on our validation set, using our top model, to identify challenges with our DeiT and ARBERT models. 
To this end, we randomly select 200 images from the validation set, divided among two experienced annotators (authors). Annotators reviewed these images, tested the model, and noted its performance. From this small-scale heuristic analysis, we identify two significant challenges our models faced:

\paragraph{Encoder-Related Issues.} Despite its effectiveness across tasks, DeiT has limitations with scalability and high-resolution inputs. It struggles with larger input sizes, impacting performance in scenarios like detailed document analysis or high-resolution character recognition. As shown in Figure \ref{fig:high_res_example}, DeiT, after undergoing unsupervised training with Masked Imaging Modeling (MIM) \cite{xie2022simmim}, may fail to reconstruct high-resolution image content due to inefficient feature processing. Conversely, SwinV2 exhibits superior handling of high-resolution inputs.

\begin{figure}[htp]
\centering
\includegraphics[width=\columnwidth]{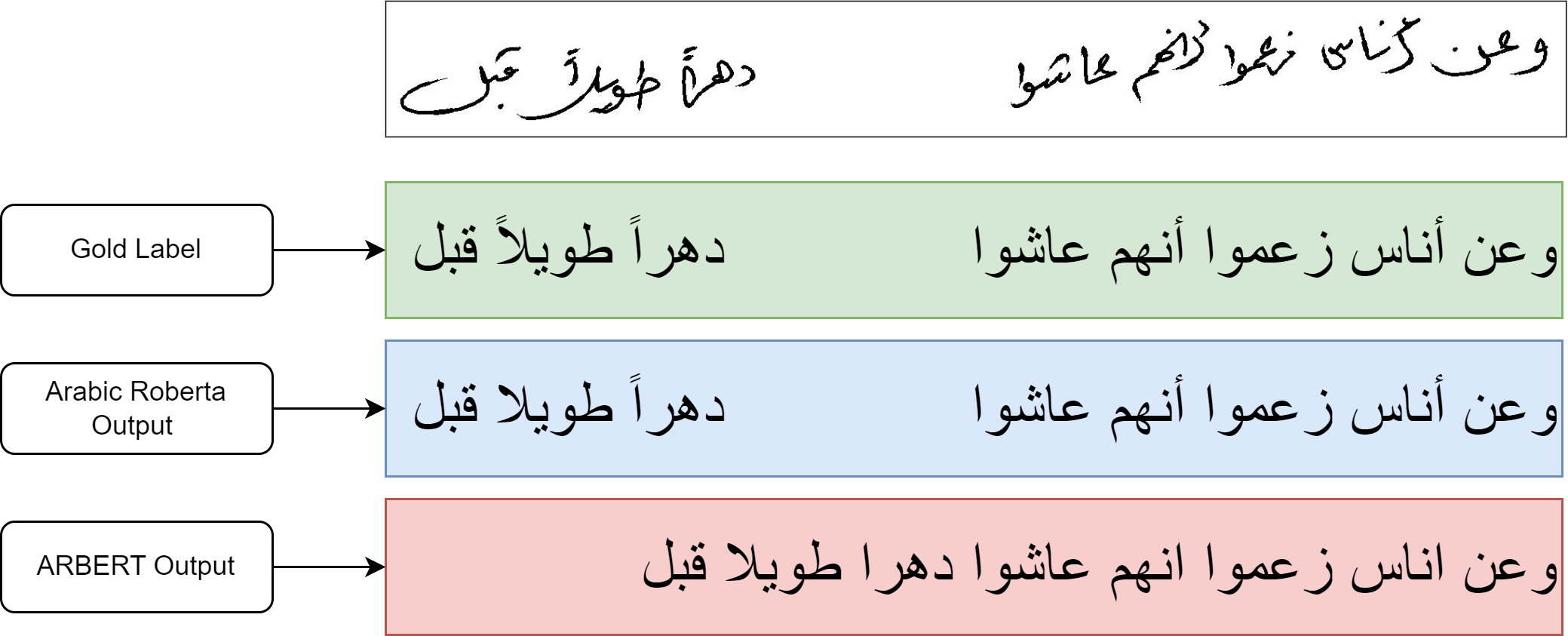}
\caption{Example of a word with diacritics that ARBERT may struggle to interpret correctly.}
\label{fig:diacritics_example}
\end{figure}

\paragraph{Decoder-Related Issues.} Regarding the decoder, ARBERT has shown limitations in handling diacritics as these are not included in its vocabulary. Diacritics play a crucial role in many languages, including Arabic and Persian, as they can significantly change word meanings. The inability of ARBERT to recognize and process these diacritics may lead to incorrect word recognition and, subsequently, incorrect text interpretation. For instance, as shown in Figure \ref{fig:diacritics_example}, ARBERT might interpret the word incorrectly due to its inability to handle diacritics.

\section{Building \textit{Qalam}} \label{qalam}

\begin{figure}
  \centering
  \includegraphics[width=\columnwidth]{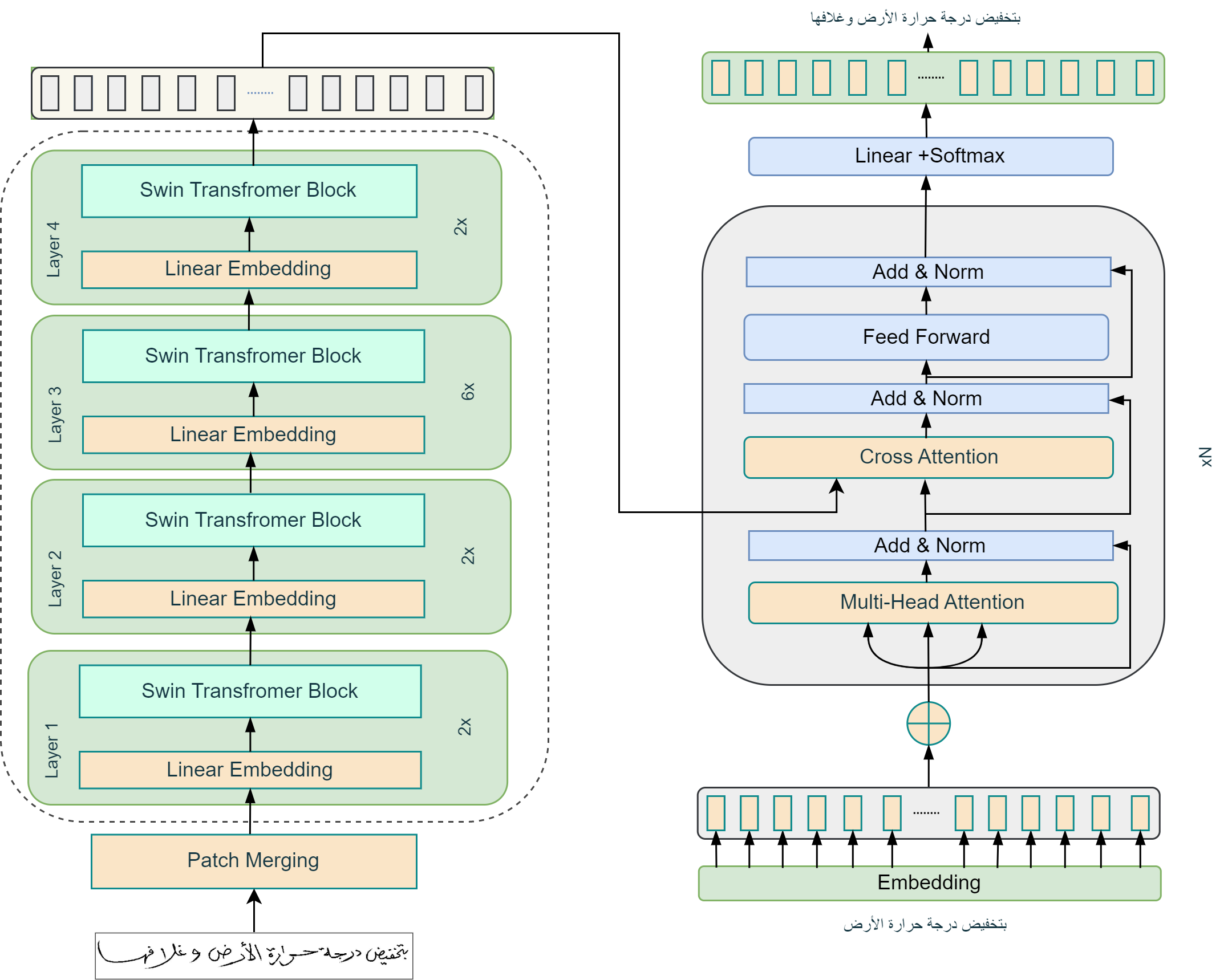}
  \caption{The architecture of \textit{Qalam} model, comprising the SwinV2 encoder for image representation and the Roberta decoder for sequence prediction, indicating the flow of data during the OCR and HWR tasks}
  \label{fig:diagram_overall}
\end{figure}

To further optimize the performance of our Vision Encoder-Decoder framework, we introduce additional pre-training strategies for the encoder and the decoder, including the SwinV2 and RoBERTa models. This is undertaken to capitalize on their specific architectural strengths in handling large image sizes and next-token prediction tasks.

\paragraph{Encoder Upgrades.} To enhance the model's performance, we further pre-train the Swin Transformer v2 encoder. We choose this model due to its proficiency in handling high-resolution inputs and capturing rich spatial information. The training strategy involved augmenting the input image size and utilizing a robust dataset comprising $4.5M$ images extracted from Arabic manuscripts and books. The Masked Language Modeling approach is employed during training, promoting the encoder's adaptability to real-world OCR challenges and the complexities inherent in Arabic script.
\paragraph{Decoder Upgrades.} On the decoder side, we further pre-train the the RoBERTa model using the Masked Language Modeling approach \cite{devlin2018bert}. We select RoBERTa as the decoder for \textit{Qalam} for its superior performance in next-token prediction tasks. A substantial and diverse training dataset provided the model with various language patterns, including Arabic Wikipedia and AraC4~\cite{abdul2023jasmine}. The model was also modified to handle longer sequences during training, improving its comprehension and generation of complex sentence structures. We used a sentence-piece tokenizer that also can understand diacritics, which is crucial for processing the Arabic language, thereby enhancing the decoder's effectiveness in Arabic OCR tasks. Figure \ref{fig:diagram_overall} shows our final architecture.

\paragraph{Synthetic Data.}  \label{synthetic}
 This data comprises $60k$ image-text pairs and features more than $28$ diverse fonts, comprehensively representing potential inputs. \textit{Qalam} is subsequently fine-tuned using this synthetic data in a supervised manner. Figure~\ref{fig:three graphs} shows a few examples of this dataset. We generate this dataset using text files from Hindawi Arabic books \cite{Hindawi}. We subsequently augment it into different fonts and organize it into PDF files of size 760x640. The model was then fine-tuned on this dataset to teach it about the Arabic script.

\begin{figure}[htbp]
     \centering
     \
     \begin{subfigure}[b]{0.4\textwidth}
         \centering
         \includegraphics[width=\textwidth]{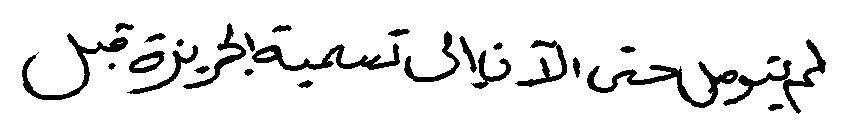}
          \caption{Real sample from the training data.}
     \end{subfigure}
     \hfill
     \begin{subfigure}[b]{0.4\textwidth}
         \centering
         \includegraphics[width=\textwidth]{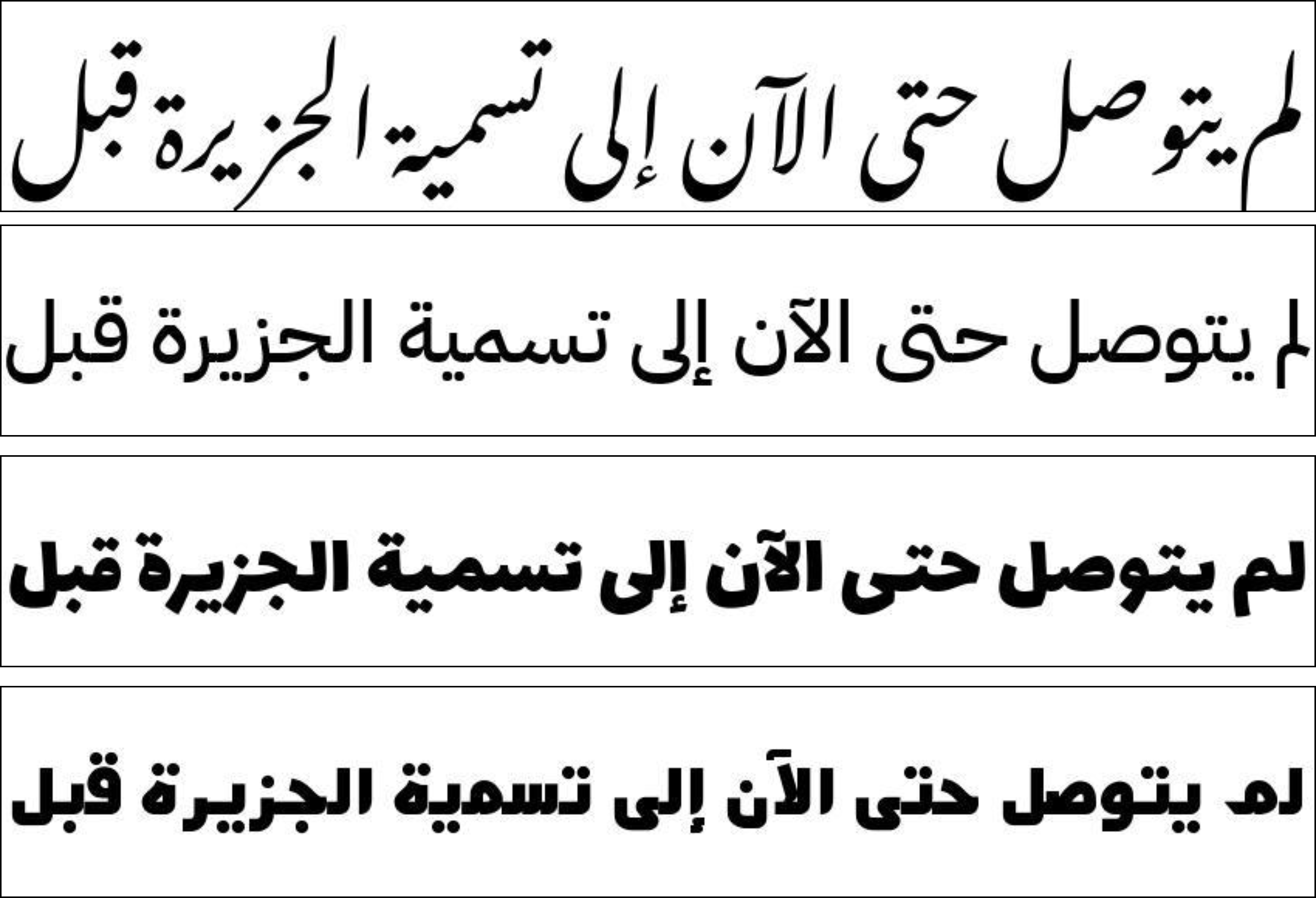}
         \caption{Augmented samples for the above sample.}
     \end{subfigure}
     
     \hfill
     
        \caption{Data augmentation for the training data.}
        \label{aug_1}
\end{figure}

\paragraph{Data Augmentation.} We supplement each training sample from the line-based datasets with four additional synthesized samples, as illustrated in Figure \ref{aug_1}. Then, we fine-tune \textit{Qalam} using this synthetic data in a supervised manner.
This strategy facilitates the model's adaptability to various text styles and complexities in Arabic scripts.
Here, we take the gold labels of the dataset's training split and augment it with four different randomly chosen fonts.

\paragraph{Training Procedure.} The training process for Qalam utilizes specific hyperparameters to optimize performance. Table~\ref{tab:hyperparameters} provides a comprehensive overview of these settings.
The model is trained using the Adam optimizer with a $5 \times 10^{-5}$ learning rate. We employ a cosine learning rate scheduler over 50 epochs. The training process uses a batch size of 8, with gradient accumulation steps of eight, resulting in a total effective batch size of 64. For reproducibility, we set the random seed to 42. The evaluation batch size is also set to eight. These carefully chosen hyperparameters are instrumental in achieving optimal performance for the Qalam model.

\begin{table*}[!htp]\centering
\scriptsize
\resizebox{\textwidth}{!}{%
\begin{tabular}{lllccccccclc}\toprule
\textbf{Cluster} &\textbf{Task} &\textbf{Dataset} &\textbf{\colorbox{green!25}{M1}} &\textbf{\colorbox{red!25}{M2}} &\textbf{\colorbox{blue!25}{M3}} &\textbf{\colorbox{yellow!25}{M4}} &\textbf{\colorbox{purple!25}{M5}} &\textbf{\colorbox{orange!25}{M6}} &\textbf{\colorbox{pink!25}{SOTA}} &\textbf{\colorbox{pink!25}{SOTA References}} &\protect\includegraphics[height=.6cm]{figs/qalam\_logo\_v2.png} \\\midrule
\multirow{5}{*}{HWR} &\multirow{2}{*}{Char} &MADBase &$124.05$ &$02.20$ &$08.00$ &$03.77$ &$00.49$ &$00.43$ &$00.57$ &\newcite{PalatnikdeSousa2018ConvolutionalEF} &\textbf{0.004} \\
& &AHCD &$109.00$ &$23.40$ &$05.43$ &$03.43$ &$01.00$ &$00.90$ &$01.58$ &\newcite{PalatnikdeSousa2018ConvolutionalEF} &\textbf{0.003} \\
&\multirow{2}{*}{Word} &ADAB &$66.00$ &$02.60$ &$12.58$ &$07.43$ &$00.99$ &$00.99$ &$01.01$ &\newcite{7490154} &\textbf{0.01} \\
& &Alexuw &$101.80$ &$05.00$ &$15.74$ &$03.23$ &$01.00$ &$00.83$ &$07.84$ &\newcite{hussein2014alexuword} &\textbf{0.01} \\
&Line &OnlineKHATT &$62.00$ &$64.32$ &$25.90$ &$15.78$ &$45.00$ &$21.54$ &$12.24$ &\newcite{OnlineKhatt} &\textbf{3.95} \\\hline
\multirow{3}{*}{OCR} &Line &PATS01 &$63.00$ &$30.23$ &$32.43$ &$26.65$ &$18.00$ &$05.23$ &n/a &- &\textbf{1.90} \\
&Word &Shotor &$124.00$ &$05.40$ &$08.32$ &$09.01$ &$02.03$ &$02.02$ &n/a &- &\textbf{0.12} \\
&Line &IDPL-PFOD &$36.00$ &$43.01$ &$29.80$ &$32.44$ &$20.00$ &$04.54$ &n/a &- &\textbf{1.53} \\\hline
\multirow{3}{*}{Overall} &\multirow{3}{*}{Avg.} &HWR\textsubscript{Score} &$92.57$ &$19.50$ &$13.53$ &$06.73$ &$09.70$ &$04.94$ &$04.65$ &- &\textbf{0.80} \\
& &OCR\textsubscript{Score} &$74.33$ &$26.21$ &$23.52$ &$22.70$ &$13.34$ &$03.93$ &n/a &- &\textbf{1.18} \\
& & \colorbox{yellow}{\textbf{\textit{\midad~\textsubscript{Score}}}} &$85.73$ &$22.02$ &$17.28$ &$12.72$ &$11.06$ &$04.56$ &$04.65$ &- & $\textbf{0.94}$ \\
\bottomrule
\end{tabular}}
\caption{Comparative performance analysis of various models across diverse OCR and HWR datasets. The models are: \colorbox{green!25}{M1}: Tesseract, \colorbox{red!25}{M2}: TrOCR Base, \colorbox{blue!25}{M3}: CRNN+CTC, \colorbox{yellow!25}{M4}:CNN+BiLSTM+CTC, \colorbox{purple!25}{M5}:DeIT+ARBERT, \colorbox{orange!25}{M6}:SwinV2(fp) + ARBERT, \colorbox{pink!25}{M7}:SOTA, and \textit{Qalam}\protect\includegraphics[height=.5cm]{figs/qalam_logo_v2.png} . The table lists WER (Lower is better) achieved by each model on different Arabic and Persian datasets classified by base, type, and language. The table also provides OCR, HWR, and \textit{MIDAD} scores to showcase the models' overall performance in respective tasks. \textit{MIDAD} Score is the average WER of across all tasks and datasets}\label{tab: final_table}
\end{table*}


\subsection{Performance Evaluation of \textit{Qalam}}
This section encapsulates the performance evaluation of various models, emphasizing our proposed model, \textit{Qalam}, alongside baseline models and alternative architectures. The results are collated in Table \ref{tab: final_table}, highlighting the exemplary performance of \textit{Qalam} across diverse datasets.

On the Handwriting Recognition (HWR) front, \textit{Qalam} exhibits remarkable performance. Specifically, in the MADBase and AHCD datasets, \textit{Qalam} can recognize all test samples without errors. Meanwhile, in word-based datasets such as ADAB and Alexuw, \textit{Qalam} achieves an equally impressive WER of just 0.01

When assessing line recognition tasks, especially on the OnlineKHATT dataset, \textit{Qalam} continues to excel, recording a WER of 3.95\%. This excellence extends to OCR tasks as well, where \textit{Qalam} upholds a low WER on various datasets, including 1.90\% on PATS01, 0.12\% on Shotor, and 1.53\% on IDPL-PFOD. These results also show that our model can perform well on Persian datasets. 

In summary, \textit{Qalam} delivers unparalleled performance across the board. The average scores indicate its superiority at a WER of 0.80\% for HWR tasks and 1.18\%  for OCR tasks. Further, it achieves a unique evaluation metric, the \textit{Midad}\textsubscript{Score}, of 0.94\%. This collective evidence positions \textit{Qalam} as a leading solution in both the HWR and OCR domains, reflecting its robustness and adaptability to diverse textual challenges.

\subsection{Performance Evaluation of \textit{Qalam} in the wild Arabic data}

\begin{table}[!htp]
\resizebox{\linewidth}{!}{%
\begin{tabular}{lrrr}
\toprule
\textbf{Datasets}  & \textbf{SOTA Ref} & \textbf{SOTA } & \multicolumn{1}{c}{\protect\includegraphics[height=.6cm]{figs/qalam\_logo\_v2.png}} \\
\midrule
KHATT & \cite{momeni2023transformer} & 18.45 & \textbf{10.43} \\
Historical Manuscripts & \cite{clausner2018icfhr} & \textbf{21.9} & 30.88 \\
\bottomrule
\end{tabular}}
\caption{ Zero Shot Evaluation of Qalam on In the wild Arabic OCR datasets.} \label{tab:comparison_qalam}
\end{table}

Table~\ref{tab:comparison_qalam} showcases the comparative performance of the zero-shot evaluation of the Qalam system on "in the wild" Arabic OCR datasets in terms of Character Error Rate (CER) as these datasets are more complex and are completely unseen by the model. The table contrasts the results of the state-of-the-art (SoTA) references with the outcomes achieved using the Qalam system. We observe the following:

\begin{itemize}
    \item \textbf{KHATT:} The SoTA result, as referenced by \citealt{momeni2023transformer}, yields a CER of $18.45$. In comparison, the Qalam system significantly improved, achieving a CER of $10.43$.
    \item \textbf{Historical Manuscripts:} The performance on this dataset provides a different scenario. The referenced SoTA result from \citealt{clausner2018icfhr} reports a CER of $21.9$. However, the Qalam system shows a higher CER of $30.88$ in this context.
\end{itemize}

In summary, while the Qalam model exhibits superior performance on the KHATT dataset, its performance on the Historical Manuscripts dataset is less competitive. The Historical Manuscripts dataset is mainly out-of-domain data, but the performance is still competitive. 

\section{Discussion}

\noindent The exemplary performance of \textit{Qalam} across Arabic and Persian OCR and HWR tasks (Table \ref{tab: final_table}) highlights its potential. Despite the diversity in the OnlineKHATT dataset, \textit{Qalam} achieves a relatively low residual error of 4\%, indicating scope for improved handling of diverse writing styles.
The superior performance of \textit{Qalam} over CTC-based models like CRNN+CTC and CNN+BiLSTM+CTC, which emphasizes the transformative potential of transformer-based models when supplemented with substantial training datasets.

The stark disparity between \textit{Qalam} and TrOCR, despite TrOCR's strength, underscores the limitation of models pretrained on English data when applied to different scripts, emphasizing the need for script-specific training.
The results also highlight the limitations of task-specific models like Tesseract, which excels in line-based recognition but underperforms in character or word-based tasks.
Finally, the exceptional performance of \textit{Qalam} can be attributed to the synergy between SwinV2 encoder and RoBERTa decoder, effectively tackling OCR and HWR complexities. Examples of the Demo can be found in the Appendix \ref{app:demo}. \\

\section{Conclusion} \label{conc}
We introduced \textit{Qalam}, a foundation model for Arabic OCR and HWR.  \textit{Qalam} establishes a new standard in Arabic OCR and HWR tasks. The robustness of its architecture, comprising the SwinV2 encoder and RoBERTa decoder, outperforms previous state-of-the-art systems. Our study demonstrates that Arabic script's unique challenges can be effectively addressed by leveraging the strengths of transformer-based models. The performance of \textit{Qalam} on the \midad~benchmark validates the scalability and flexibility of our approach, suggesting its potential application to OCR and HWR tasks in other complex scripts. Moving forward, \textit{Qalam} offers a compelling basis for further innovation in Arabic OCR and HWR systems, contributing to the advancement of this critical area of research.

\section {Limitations} \label{lim}
Given the limited availability of HWR and OCR datasets for Arabic, particularly for handwriting with diacritics - a feature often omitted in everyday writing - certain challenges arise. The most notable of these is the prevalence of code-switching and dialectal in real-world writing, both in OCR and HWR contexts, which the current model may struggle to address. Complex tasks such as Scene Text Recognition (STR), multiline, and full-page recognition also show significant limitations to the capabilities of \textit{Qalam}. 

Furthermore, \textit{Qalam} has been specifically designed for Arabic OCR and HWR tasks. As a result, its performance has not been assessed on other scripts or languages. Therefore, its effectiveness in these contexts may not be optimal without further modifications and fine-tuning. This should be considered when attempting to generalize \textit{Qalam}'s capabilities beyond Arabic OCR and HWR tasks.

\section*{Acknowledgments}\label{sec:acknow}
We acknowledge support from Canada Research Chairs (CRC), the Natural Sciences and Engineering Research Council of Canada (NSERC; RGPIN-2018-04267), the Social Sciences and Humanities Research Council of Canada (SSHRC; 435-2018-0576; 895-2020-1004; 895-2021-1008), Canadian Foundation for Innovation (CFI; 37771), Digital Research Alliance of Canada,\footnote{\href{https://alliancecan.ca}{https://alliancecan.ca}} and UBC ARC-Sockeye.




\bibliography{anthology,custom}

\appendix


\section{Appendices} \label{appendices}

We provide an addition organized as follows:
\begin{itemize}
\item Model architectures in Section \ref{models_arch}.
\item Datasets Details  Section \ref{details}.
\item WER equation Section \ref{wer_eq}. 
\item Hyperparameter Table Section \ref{hyp_tab}.
\item Synthetic Data \ref{Synth_dog}
\item \textit{Qalam} demo Section \ref{app:demo}.
\item Test Results \ref{tab: test}
\end{itemize}

\subsection{Model architectures in the literature } \label{models_arch}
\begin{figure*}[htbp]
    \centering
    \resizebox*{.75\linewidth}{!}{%
        \small{ 
            \begin{forest}
                forked edges,
                for tree={
                    grow'=east,
                    draw,
                    rounded corners,
                    text width=5.5cm,
                    node options={align=center},
                }       
                [Models, fill=col0, parent, s sep=1cm, node options={align=center, rotate=0}
                    [HMMs \\ \cite{agazzi1993hidden,bunke1995off,park1996off,alma2002recognition,prasad2008improvements} ,text width=3.5cm,fill=col1
                    ]
                    [CTC-based\\ \cite{graves2006connectionist,graves2008offline} , for tree={child, fill=col2}
                        [RNN \\
                        \cite{pham2014dropout,su2015accurate,ahmad2017khatt}
                        ,text width=6cm]
                        [CNN-RNN \\ 
                        \cite{bluche2017gated,breuel2017high,de2020htr,shi2016end}
                        ,text width=6cm]
                        [CNNs \\ \cite{coquenet2020recurrence,yousef2020accurate}
                        ,text width=6cm , fill=col2]
                    ] 
                    [Encoder-Decoder \\  \cite{sutskever2014sequence,bluche2017scan}, for tree={child, fill=col3}
                        [RNN \\
                        \cite{voigtlaender2016handwriting,bluche2017scan}
                        ,text width=6cm]
                        [CNN-RNN \\
                        \cite{lee2016recursive,sueiras2018offline,shi2018aster}
                        ,text width=6cm]
                        [RNN with Attention \\ \cite{doetsch2016bidirectional,coquenet2022end}
                        ,text width=6cm ]
                        [Transfromer \\ 
                        \cite{vaswani2017attention}, for tree={child}
                            [From Scratch \\
                            \citet{li2021trocr,kang2022pay,barrere2022light,wick2022rescoring}
                            ,text width=6cm]
                            [Pre-trained \\ 
                            \cite{devlin2018bert,mostafa2021ocformer,kim2022ocr,lyu2022maskocr,momeni2023transformer}
                            ,text width=6cm]
                        ]
                    ]
                ]
            \end{forest}
        }
    }
    \caption{A categorization of diverse model architectures leveraged in the literature, providing an overarching view of the methodological landscape in OCR and HWR.}
    \label{fig:classification}
\end{figure*}
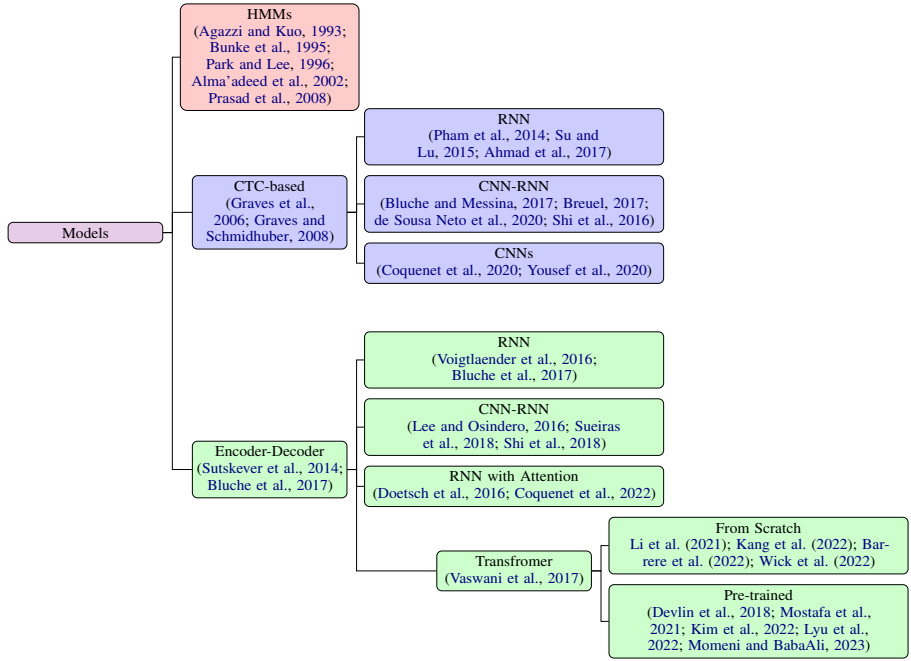

In this section, we provide an illustrative Figure \ref{fig:classification} of various model architectures used in the literature.

\subsection{Dataset Details} \label{details}
Table \ref{tab:datasets}  provides additional statistics  \midad.

\subsection{WER Equation}\label{wer_eq}
In this section, we present the equation used for the calculation of WER:
\begin{equation} \label{eqn}
    WER = \frac{(S + D + I)}{N} = \frac{(S + D + I)}{(S + D + C)}
\end{equation}
where:
\begin{equation*}
\begin{aligned}
S & : \text{number of substitutions}, \\
D & : \text{number of deletions}, \\
I & : \text{number of insertions}, \\
C & : \text{number of correct words}, \\
N & : \text{number of words in the reference}.
\end{aligned}
\end{equation*}

\subsection{Hyperparameter Table}    \label{hyp_tab}
In this section, we present Table \ref{tab:hyperparameters}, detailing the hyperparameters employed in our study.
\begin{table}[!htp]\centering
\scriptsize
\resizebox{\linewidth}{!}{%
\begin{tabular}{lc}\toprule
\textbf{Hyperparameter} &\textbf{Value} \\\toprule
Learning Rate & $5 \times 10^{-5}$ \\
Train Batch Size & 8 \\
Eval Batch Size & 8 \\
Seed & 42 \\
Gradient Accumulation Steps & 8 \\
Total Train Batch Size & 64 \\
Optimizer & Adam (betas=(0.9,0.999), epsilon=$1 \times 10^{-8}$) \\
LR Scheduler Type & Cosine \\
Num Epochs & 50 \\ \bottomrule
\end{tabular}}
\caption{Summary of hyperparameters used for the training process.}\label{tab:hyperparameters}
\end{table}

\subsection{Synthetic Data} \label{Synth_dog}
\begin{figure}[htbp]
     \centering
     \begin{subfigure}[b]{0.35\textwidth}
         \centering
         \includegraphics[width=\textwidth]{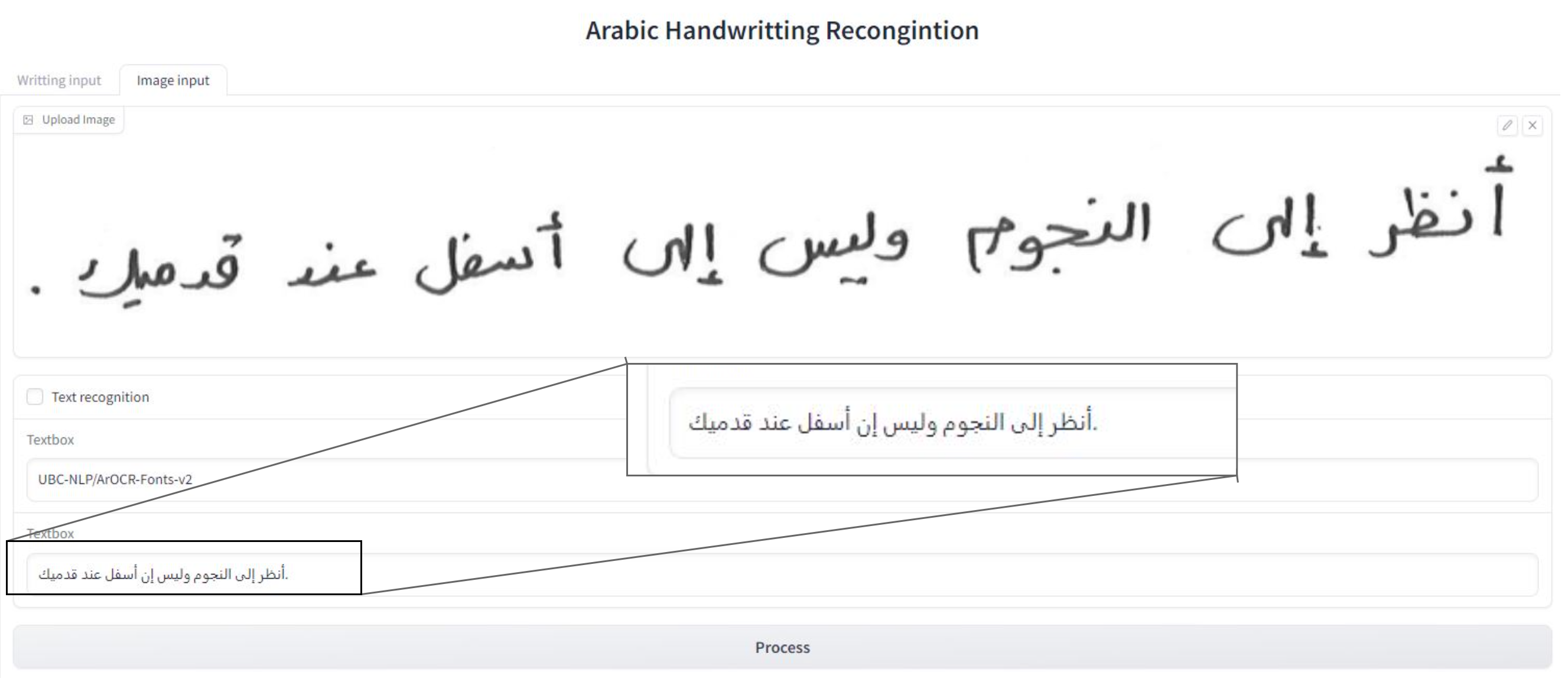}
     \end{subfigure}
     \hfill
     \begin{subfigure}[b]{0.4\textwidth}
         \centering
         \includegraphics[width=\textwidth]{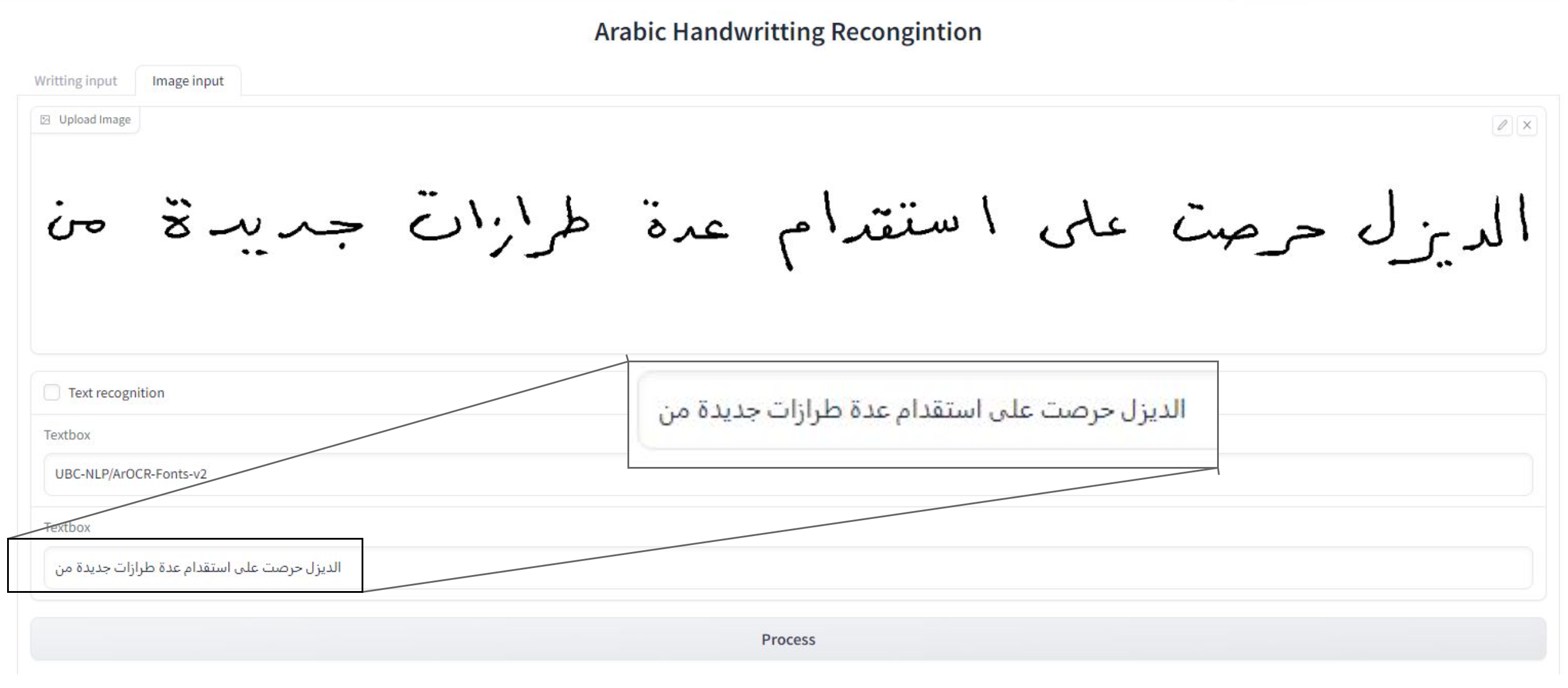}
     \end{subfigure}
          \hfill
      \begin{subfigure}[b]{0.4\textwidth}
         \centering
         \includegraphics[width=\textwidth]{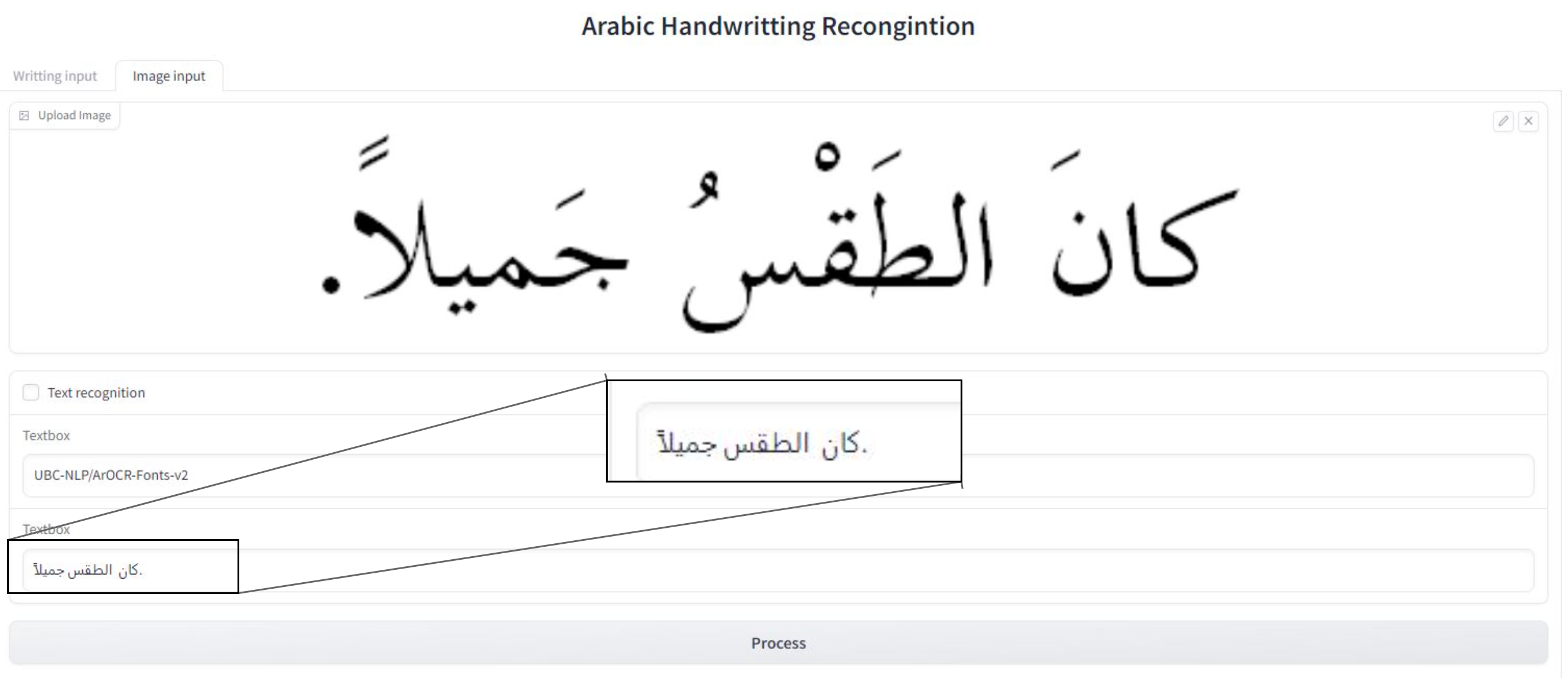}
     \end{subfigure}
          \hfill
      \begin{subfigure}[b]{0.4\textwidth}
         \centering
         \includegraphics[width=\textwidth]{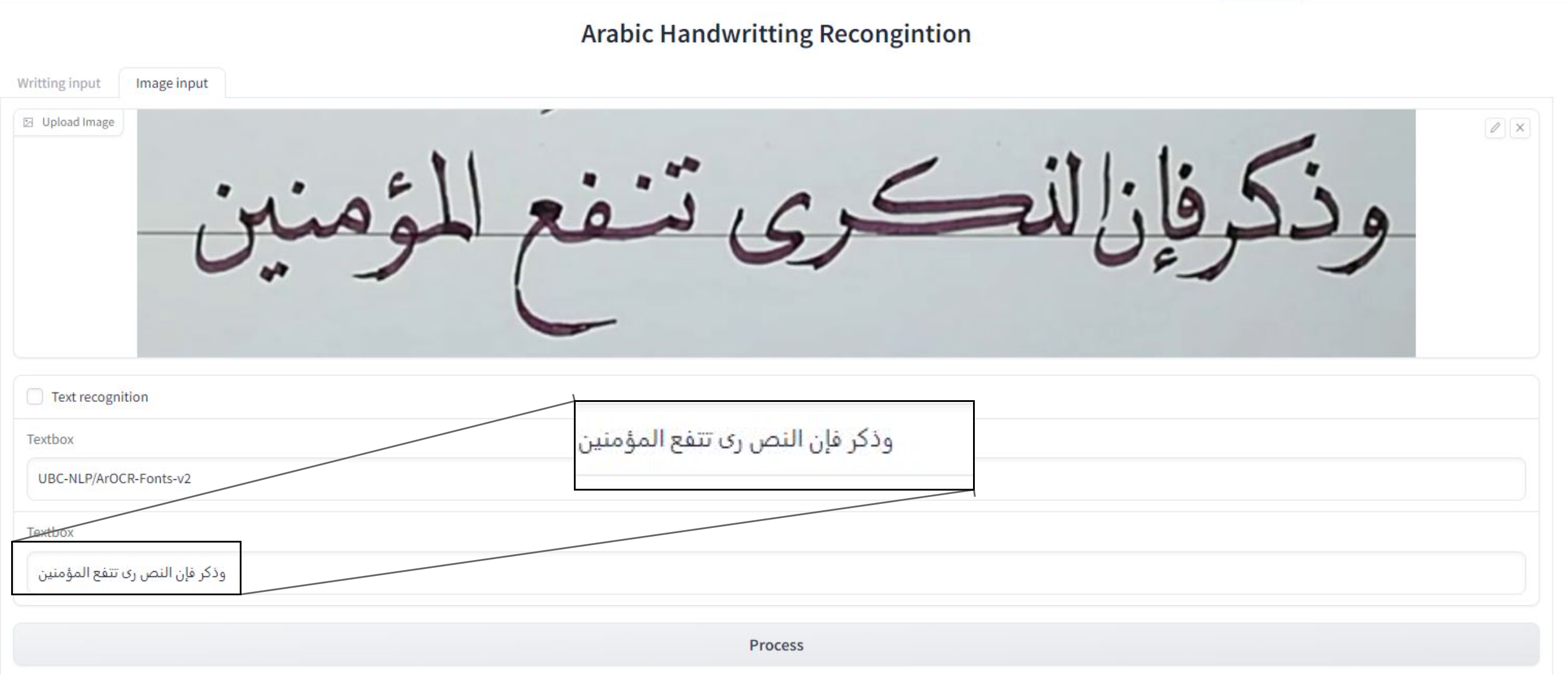}
     \end{subfigure}
     \hfill
     
        \caption{\textit{Qalam} Demo samples.}
        \label{fig:three graphs}
\end{figure}

\subsection{\textit{Qalam} Demo}  \label{app:demo}
In addition to the computational experiments, we also developed a practical demonstration that accepts two types of inputs: handwriting and images. The handwriting input facilitates our model's HWR capabilities, allowing users to test the system's performance in real-time. Simultaneously, the image input caters to OCR tasks, enabling users to upload images of Arabic scripts and observe the model's interpretation. 
One of the noteworthy features of our model is its capacity to handle complex diacritics, a characteristic intrinsic to Arabic scripts. Arabic diacritics are essential in the language, affecting word meanings and pronunciations. However, their tiny size and positioning above or below the line of text make them challenging for many OCR systems. As evidenced by the demonstration, our model exhibits robust performance in recognizing and interpreting these diacritics. 
The proficiency of our model isn't limited to diacritics; it extends to handling various types of Arabic texts. Whether it be different fonts, styles, or levels of complexity, our system's adaptability makes it a potent tool for Arabic script recognition. The demonstration provides a tangible testament to these capabilities, illustrating how the advancements in our model translate into practical, real-world applications. Additionally, Figure \ref{fig:three graphs} displays screenshots of some synthetic samples used in our study.

     

\subsection{Test Results}\label{tab: test}

\begin{table}[!htp]\centering
 \scriptsize
\resizebox{\linewidth}{!}{%
\begin{tabular}{lrrrrrrrr}\toprule
\textbf{Cluster } &\textbf{Task } &\textbf{Dataset} &\textbf{\colorbox{green!25}{E1}} &\textbf{\colorbox{red!25}{E2}} &\textbf{\colorbox{orange!25}{E3}} &\textbf{\colorbox{blue!25}{E4}} &\textbf{\colorbox{purple!25}{E5}} \\\midrule
\multirow{5}{*}{\textbf{HWR}} &\multirow{2}{*}{Char} &MADBase &5.99 &4.89 &5.34 &4.90 &\textbf{4.24} \\
& &AHCD &8.49 &10.34 &7.45 &6.71 &\textbf{5.23} \\
&\multirow{2}{*}{Word} &ADAB &5.42 &4.52 &9.42 &4.55 &\textbf{4.35} \\
& &Alexuw &12.42 &9.57 &9.28 &9.42 &\textbf{6.28} \\
&Line &OnlineKHATT &73.52 &67.89 &66.73 &\textbf{65.60} &66.53 \\\midrule
\multirow{3}{*}{\textbf{OCR}} &Line &PATS01 &51.23 &39.24 &43.23 &40.76 &\textbf{32.21} \\
&Word &Shotor &10.24 &9.80 &9.23 &9.16 &\textbf{8.53} \\
&Line &IDPL-PFOD &45.34 &49.93 &32.89 &31.70 &\textbf{30.92} \\\midrule
\multirow{3}{*}{\textbf{Overall}} &\multirow{3}{*}{\textbf{Avg.}} &HWR Score &21.17 &19.44 &19.65 &21.99 &\textbf{17.33} \\
& &OCR Score &35.61 &32.99 &28.45 &27.21 &\textbf{23.89} \\
& &\colorbox{yellow}{\textbf{\textit{Midad~\textsubscript{Score}}}} &26.58 &24.52 &22.95 &21.60 &\textbf{19.79} \\
\bottomrule
\end{tabular}}
\caption{Comparative performance analysis of various models across diverse OCR and HWR datasets. \colorbox{green!25}{E1}: ViT, \colorbox{red!25}{E2}: Swin, \colorbox{orange!25}{E3}: BeiT, 
\colorbox{blue!25}{E4}: SwinV2, 
and \colorbox{purple!25}{E5}:DeiT with XLM-R as decoder. The table also provides OCR, HWR, and \textit{Midad} scores to showcase the models' overall performance in respective tasks.}\label{tab: encoder_table}
\end{table}

\begin{table}[!htp]\centering
\scriptsize
\resizebox{\linewidth}{!}{%
\begin{tabular}{lrrrrrrrr}\toprule
\textbf{Cluster } &\textbf{Task } &\textbf{Dataset} &\textbf{\colorbox{green!25}{D1}} &\textbf{\colorbox{red!25}{D2}} &\textbf{\colorbox{orange!25}{D3}} &\textbf{\colorbox{blue!25}{D4}} &\textbf{\colorbox{purple!25}{D5}} \\\midrule
\multirow{5}{*}{\textbf{HWR}} &\multirow{2}{*}{Char} &MADBase  &$12.53$ &4.24 &5.53 &2.65 &\textbf{0.53} \\
& &AHCD  &13.99 &5.23 &9.10 &2.31 &\textbf{1.35} \\
&\multirow{2}{*}{Word} &ADAB &9.01 &4.35 &6.07 &3.53 &\textbf{1.63} \\
& &Alexuw &13.42 &6.28 &9.24 &5.93 &\textbf{1.43} \\
&Line &OnlineKHATT &74.76 &66.53 &67.09 &59.23 &\textbf{55.34} \\\midrule
\multirow{3}{*}{\textbf{OCR}} &Line &PATS01 &53.53 &32.21 &45.11 &28.08 &\textbf{23.42} \\
&Word &Shotor &10.89 &8.53 &6.54 &6.01 &\textbf{4.12} \\
&Line &IDPL-PFOD &64.10 &30.92 &35.65 &\textbf{28.42} &28.95 \\\midrule
\multirow{3}{*}{\textbf{HWR+OCR}}&\multirow{3}{*}{\textbf{Avg.}}&HWR Score &24.74 &17.33 &19.41 &14.73 &\textbf{12.06} \\
& &OCR Score &42.84 &23.89 &29.10 &20.84 &\textbf{18.83} \\
& &\colorbox{yellow}{\textbf{\textit{\midad~\textsubscript{Score}}}} &31.53 &19.79 &23.04 &17.02 &\textbf{14.60} \\
\bottomrule 
\end{tabular}}
\caption{Performance comparison of various transformer-based decoder models on HWR and OCR tasks for different datasets. \colorbox{green!25}{D1}: RoBERTa, \colorbox{red!25}{D2}: XLM-R, \colorbox{orange!25}{D3}: MARBERT, \colorbox{blue!25}{D4}: MARBERT\textsubscript{v2}, and \colorbox{purple!25}{D5}: ARBERT with DeiT as constant encoder. Tasks are categorized into character-level (Char), word-level (Word), and line-level (Line) recognition.}\label{tab: decoder_table}
\end{table}




\end{document}